\documentclass[pdflatex,sn-mathphys-num]{sn-jnl}

\usepackage{graphicx}%

\usepackage{amsmath,amssymb,amsfonts}%
\usepackage{amsthm}%
\usepackage{mathrsfs}%
\usepackage{tabularray}
\usepackage[title]{appendix}%
\usepackage{xcolor}%
\usepackage{textcomp}%
\usepackage{manyfoot}%
\usepackage{booktabs}%
\usepackage{algorithm}%
\usepackage{algorithmicx}%
\usepackage{algpseudocode}%
\usepackage{listings}%
\usepackage{xcolor}%
\usepackage{textcomp}%
\usepackage{manyfoot}%
\usepackage{booktabs}%
\usepackage{algorithm}%
\usepackage{multicol}
\usepackage{multirow}
\usepackage{algorithmicx}%
\usepackage{algpseudocode}%
\usepackage{listings}%
\usepackage{rotating}
\usepackage{booktabs}
\usepackage{makecell}
\usepackage{float}
\usepackage{adjustbox}
\usepackage{pdflscape}
\usepackage{alphalph}

\theoremstyle{thmstyleone}%
\theoremstyle{thmstyletwo}%

\theoremstyle{thmstylethree}%

\begin{document}

\title[The Foundational Capabilities of Large Language Models in Predicting Postoperative Risks Using Clinical Notes]{The Foundational Capabilities of Large Language Models in Predicting Postoperative Risks Using Clinical Notes}


\author[1,2,3,4]{\fnm{Charles} \sur{Alba}}\email{alba@wustl.edu}
\equalcont{These authors contributed equally to this work.}

\author[1,2]{\fnm{Bing} \sur{Xue}}\email{bingxue@wustl.edu}
\equalcont{These authors contributed equally to this work.}

\author[1,5,6]{\fnm{Joanna} \sur{Abraham}}\email{joannaa@wustl.edu}

\author[1,2,5,6]{\fnm{Thomas} \sur{Kannampallil}}\email{thomas.k@wustl.edu}

\author*[1,2,5]{\fnm{Chenyang} \sur{Lu}}\email{lu@wustl.edu}

\affil[1]{\orgdiv{AI for Health Institute}, \orgname{Washington University in St. Louis}, \orgaddress{\street{1 Brookings Drive}, \city{St Louis}, \postcode{63130}, \state{MO}, \country{USA}}}

\affil[2]{\orgdiv{McKelvey School of Engineering}, \orgname{Washington University in St Louis}, \orgaddress{\street{1 Brookings Drive}, \city{St Louis}, \postcode{63130}, \state{MO}, \country{USA}}}

\affil[3]{\orgname{Brown School}, \orgname{Washington University in St Louis}, \orgaddress{\street{1 Brookings Drive}, \city{St Louis}, \postcode{63130}, \state{MO}, \country{USA}}}

\affil[4]{\orgname{National University of Singapore}, \orgaddress{\street{21 Lower Kent Ridge Rd}, \city{Singapore}, \postcode{119077}}}

\affil[5]{\orgdiv{School of Medicine}, \orgname{Washington University in St Louis}, \orgaddress{\street{660 S Euclid Ave}, \city{St. Louis}, \postcode{63110}, \state{MO}, \country{USA}}}

\affil[6]{\orgdiv{Institute for Informatics, Data Science, and Biostatistics}, \orgname{Washington University in St Louis}, \orgaddress{\street{660 S Euclid Ave}, \city{St. Louis}, \postcode{63110}, \state{MO}, \country{USA}}}


\abstract{Clinical notes recorded during a patient's perioperative journey holds immense informational value. Advances in large language models (LLMs) offer opportunities for bridging this gap. Using 84,875 pre-operative notes and its associated surgical cases from 2018 to 2021, we examine the performance of LLMs in predicting six postoperative risks using various fine-tuning strategies. Pretrained LLMs outperformed traditional word embeddings by an absolute AUROC of 38.3\% and AUPRC of 33.2\%. Self-supervised fine-tuning further improved performance by 3.2\% and 1.5\%. Incorporating labels into training further increased AUROC by 1.8\% and AUPRC by 2\%. The highest performance was achieved with a unified foundation model, with improvements of 3.6\% for AUROC and 2.6\% for AUPRC compared to self-supervision, highlighting the foundational capabilities of LLMs in predicting postoperative risks, which could be potentially beneficial when deployed for perioperative care}




\maketitle

\section{Introduction}\label{sec1}



More than 10\% of surgical patients experience major postoperative complications, such as pneumonia and blood clots ranging from pulmonary embolism (PE) to deep vein thrombosis (DVT) \cite{Xue_2021b,Hamel_2005,Healey_2002, Turrentine_2006, Xue_perioperative_2022}. These complications often lead to increased mortality, intensive care unit admissions, extended hospital stays, and higher healthcare costs \cite{Tevis_2013}. Many of these preventable complications could be avoided through early identification of patient risk factors  \cite{centre2020impact,Wolfhagen_2022}. Recent reports show that clear perioperative pathways, defined as non-surgical procedures before and after surgery like shared decision-making, pre-operative assessments, enhanced surgical preparation and discharge planning, can reduce hospital stays by an average of two days across various surgeries, with subsequently carefully designed interventions reducing complications by 30\% to 80\%. This far exceeds the effects of drug or treatment interventions \cite{centre2020impact}. This underscores the critical role of identifying patient risk factors early on and implementing effective preventive measures to improve patient outcomes.

Most machine learning models aimed at predicting postoperative risks primarily utilize numerical and categorical variables, or time-series measurements \cite{Xue_2021b, Xue_perioperative_2022}. These models typically include features such as demographics, history of comorbidities, lab tests, medications, and statistical features extracted from time series \cite{Xue_2021b,Xue_perioperative_2022}, along with features reflecting surgical settings like scheduled surgery duration, surgeon name, anesthesiologist name, and location of the operating room \cite{Xue_perioperative_2022}, as well as factors such as drug dosing, blood loss, and vital signs \cite{hofer2020development}.

Text-based clinical notes taken during the surgical care journey hold immense informational value, with the potential to predict postoperative risks. Unlike discrete electronic health records (EHRs) like tabular or time-series data, clinical notes represents a form of clinical narratives, thereby allowing clinicians to communicate personalized accounts of a patients history's that could not otherwise be conveyed through traditional tabular data \cite{spasic2020clinical}. Informational value from clinical notes could therefore aid the decision-making processes that impact the course of a patient's perioperative journey and beyond \cite{Braaf_2011}. This includes the preparation for surgery, the transfer of patients to operating rooms, and the prioritization of clinicians' tasks \cite{Braaf_2011,Riley_2007}, emphasizing their importance in achieving safe patient outcomes \cite{FitzHenry,Riley_2007}. 


Following the emergence of ChatGPT, a large body of research on LLMs in medicine has been primarily centered on the application and development of conversational chatbots to support clinicians, provide patient care, and enhance clinical decision support \cite{nazi2024large}. These efforts include automating medical reports \cite{wang2023chatcad}, creating personalized treatments \cite{yang2023exploring}, and generating summaries \cite{Van_2024} or recommendations \cite{nazi2024large} from clinical notes, respectively. However, there has been comparatively less exploration of specialized fine-tuned models in directly analyzing clinical notes to make robust predictions regarding surgical outcomes and complications. Most work has primarily leveraged publicly available resources, such as open-sourced EHR databases like MIMIC to predict event-based outcomes such as ICU readmission \cite{huang2019clinicalbert} or benchmarks like i2b2 to perform information extraction tasks, like identifying concepts, diagnoses, and medications mentioned in clinical notes \cite{alsentzer2019publicly}. Some models have been developed for specialized medical diagnostics, such as COVID-19 prediction \cite{li2023text} and Alzheimer's disease (AD) progression prediction \cite{mao2023ad}. However, the application of LLMs in predicting specific postoperative surgical complications has remained limited.

\begin{figure}[h!]
    \centering
    \includegraphics[width=\textwidth]{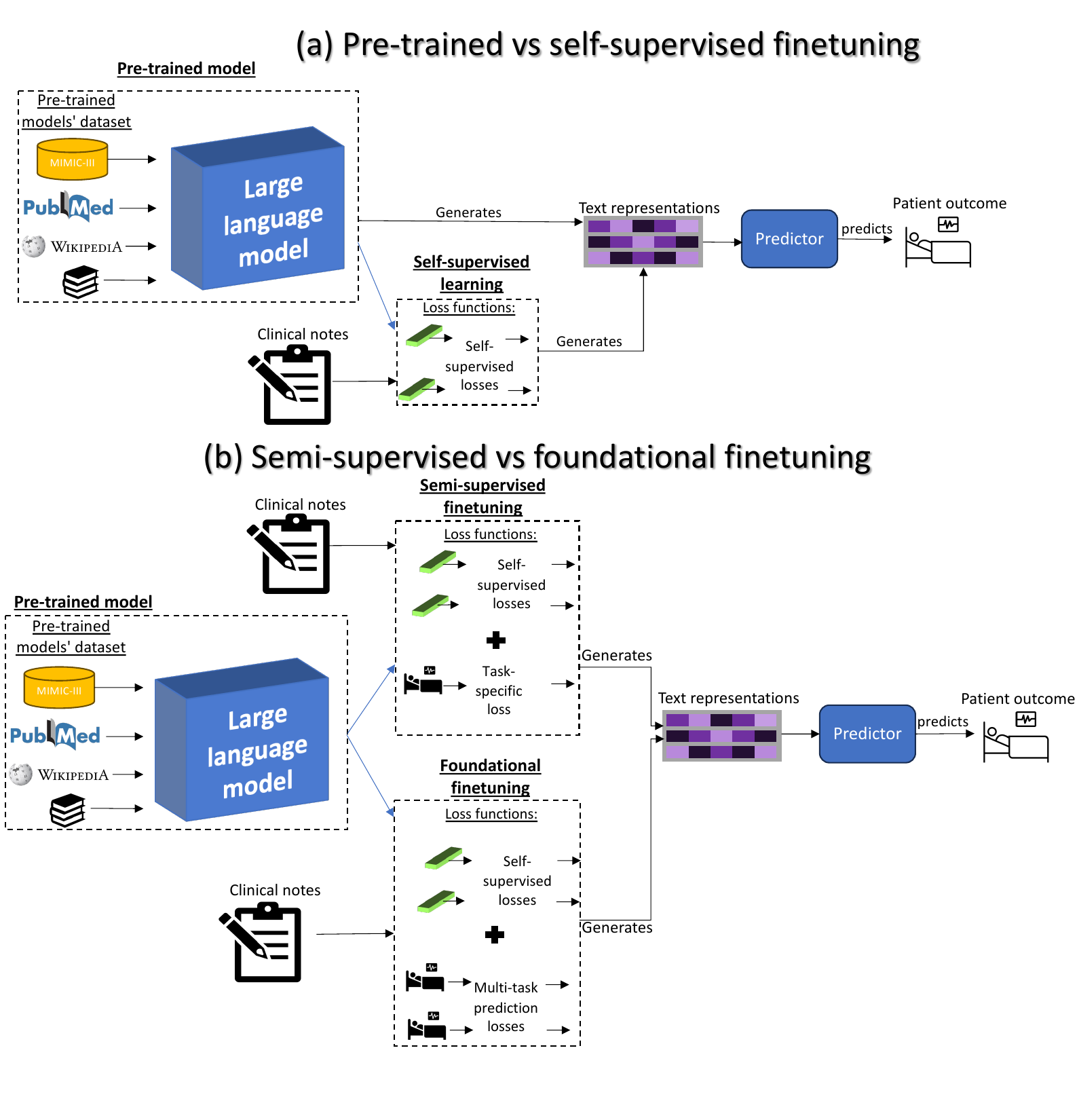}
    \caption{An illustration of the architectures encompassing different fine-tuning strategies experimented in our study, encompassing the results reported from Sections \ref{sec:pretrained_vs_word} to \ref{sec:foundational}. Fig 1a (top) illustrates how using the pretrained model alone differs from self-supervised fine-tuning when clinical texts are provided to the pretrained LLM to refine the model weights with respect to its objective loss functions. Fig 1b (below) illustrates two separate fine-tuning strategies: semi-supervised fine-tuning – creating a model that is fine-tuned under the supervision of a specific outcome; and foundation fine-tuning – creating a foundation model that is fine-tuned through a multi-task learning (MTL) objective using all available postoperative labels in the dataset.}
    \label{fig:finetuning_methods}
\end{figure}

Therefore, our work aims to develop models for the prediction of specific postoperative surgical complications through pre-operative clinical notes, with the goal of enabling the early identification of patient risk factors. By mitigating adverse surgical outcomes, we hope to leverage LLMs to enhance patient safety and improve patient outcomes. Henceforth, our contributions are multifold: First, we explore the predictive performance that pre-trained language models offer in comparison to traditional word embeddings. Specifically, by comparing the predictive performance of postoperative risk predictions between clinically-oriented pre-trained LLMs and traditional word embeddings, we demonstrate the potential of pre-trained transformer-based models in predicting postoperative complications from pre-operative notes, relative to traditional NLP methods. Second, we explore a series of finetuning techniques to enhance the performance of LLMs for predicting postoperative outcomes, as illustrated in Figure \ref{fig:finetuning_methods}. This is critical for understanding the potential of pre-trained models in perioperative care and determining optimal strategies to maximize their predictive capabilities in the early identification of postoperative risks. Third, we demonstrate the foundational capabilities of LLMs in the early identification of postoperative complications from pre-operative care notes. This suggests that a unified model could be applied directly to predict a wide range of risks, rather than training separate models for each specific clinical use case. The versatility of such foundation models could offer immense tangible benefits when deployed in real-life clinical settings.

\section{Results}\label{sec2}

A total of 84,875 clinical notes containing descriptions of scheduled procedures derived from electronic anesthesia records (Epic) of all adult patients (mean [SD] age, 56.9 [16.8] years; 50.3\% male; 74\% White) at Barnes Jewish Hospital (BJH) from 2018 to 2021 were included in this study. The six outcomes were death within 30 days (1,694 positive cases; positive event rate of 2\%), Deep Vein Thrombosis (DVT) (498 positive cases; positive event rate of 0.6\%), Pulmonary Embolism (PE) (287 positive cases; positive event rate of 0.3\%), pneumonia (475 positive cases; positive event rate of 0.6\%), Acute Kidney Injury (AKI) (11,418 positive cases; positive event rate of 13\%), and delirium (5,695 positive cases; positive event rate of 47\%).

\subsection{The state of pretrained LLMs in perioperative care}
\label{sec:pretrained_vs_word}

\begin{sidewaystable}[]

\resizebox{\textwidth}{!}{\begin{tabular}{ccccccccccccc}
\toprule
\multirow{2}{*}{Model} & \multicolumn{2}{c}{Death in 30 days}                                                                                            & \multicolumn{2}{c}{PE}                                                                                                          & \multicolumn{2}{c}{Pneumonia}                                                                                                   & \multicolumn{2}{c}{DVT}                                                                                                          & \multicolumn{2}{c}{AKI}                                                                                                         & \multicolumn{2}{c}{Delirium}                                                                                                    \\ \cmidrule{2-13} 
                       & AUROC                                                          & AUPRC                                                          & AUROC                                                          & AUPRC                                                          & AUROC                                                          & AUPRC                                                          & AUROC                                                          & AUPRC                                                           & AUROC                                                          & AUPRC                                                          & AUROC                                                          & AUPRC                                                          \\ \midrule
cbow                   & \begin{tabular}[c]{@{}c@{}}0.528\\ (0.409, 0.648)\end{tabular} & \begin{tabular}[c]{@{}c@{}}0.023\\ (0.015, 0.031)\end{tabular} & \begin{tabular}[c]{@{}c@{}}0.506\\ (0.418, 0.593)\end{tabular} & \begin{tabular}[c]{@{}c@{}}0.004\\ (0.002, 0.006)\end{tabular} & \begin{tabular}[c]{@{}c@{}}0.526\\ (0.384, 0.668)\end{tabular} & \begin{tabular}[c]{@{}c@{}}0.009\\ (0.001, 0.016)\end{tabular} & \begin{tabular}[c]{@{}c@{}}0.524 \\ (0.457, 0.59)\end{tabular} & \begin{tabular}[c]{@{}c@{}}0.006 \\ (0.005, 0.007)\end{tabular} & \begin{tabular}[c]{@{}c@{}}0.56\\ (0.488, 0.632)\end{tabular}  & \begin{tabular}[c]{@{}c@{}}0.156\\ (0.125, 0.187)\end{tabular} & \begin{tabular}[c]{@{}c@{}}0.501\\ (0.464, 0.539)\end{tabular} & \begin{tabular}[c]{@{}c@{}}0.474\\ (0.441, 0.507)\end{tabular} \\
Doc2vec                & \begin{tabular}[c]{@{}c@{}}0.479\\ (0.348, 0.611)\end{tabular} & \begin{tabular}[c]{@{}c@{}}0.021\\ (0.012, 0.03)\end{tabular}  & \begin{tabular}[c]{@{}c@{}}0.517\\ (0.466, 0.567)\end{tabular} & \begin{tabular}[c]{@{}c@{}}0.004\\ (0.004, 0.004)\end{tabular} & \begin{tabular}[c]{@{}c@{}}0.495\\ (0.347, 0.643)\end{tabular} & \begin{tabular}[c]{@{}c@{}}0.006\\ (0.003, 0.01)\end{tabular}  & \begin{tabular}[c]{@{}c@{}}0.531\\ (0.443, 0.619)\end{tabular} & \begin{tabular}[c]{@{}c@{}}0.007 \\ (0.004, 0.01)\end{tabular}  & \begin{tabular}[c]{@{}c@{}}0.523\\ (0.421, 0.624)\end{tabular} & \begin{tabular}[c]{@{}c@{}}0.146\\ (0.092, 0.199)\end{tabular} & \begin{tabular}[c]{@{}c@{}}0.484\\ (0.439, 0.53)\end{tabular}  & \begin{tabular}[c]{@{}c@{}}0.466\\ (0.417, 0.514)\end{tabular} \\
fastText               & \begin{tabular}[c]{@{}c@{}}0.725\\ (0.67, 0.781)\end{tabular}  & \begin{tabular}[c]{@{}c@{}}0.05\\ (0.04, 0.06)\end{tabular}    & \begin{tabular}[c]{@{}c@{}}0.652\\ (0.602, 0.701)\end{tabular} & \begin{tabular}[c]{@{}c@{}}0.007\\ (0.005, 0.01)\end{tabular}  & \begin{tabular}[c]{@{}c@{}}0.696\\ (0.643, 0.749)\end{tabular} & \begin{tabular}[c]{@{}c@{}}0.016\\ (0.008, 0.024)\end{tabular} & \begin{tabular}[c]{@{}c@{}}0.694\\ (0.642, 0.746)\end{tabular} & \begin{tabular}[c]{@{}c@{}}0.014\\ (0.011, 0.017)\end{tabular}  & \begin{tabular}[c]{@{}c@{}}0.726\\ (0.702, 0.75)\end{tabular}  & \begin{tabular}[c]{@{}c@{}}0.273\\ (0.239, 0.307)\end{tabular} & \begin{tabular}[c]{@{}c@{}}0.565\\ (0.541, 0.589)\end{tabular} & \begin{tabular}[c]{@{}c@{}}0.533\\ (0.513, 0.554)\end{tabular} \\
GloVe                  & \begin{tabular}[c]{@{}c@{}}\underline{0.818}\\ \underline{(0.807, 0.83)}\end{tabular}  & \begin{tabular}[c]{@{}c@{}}\underline{0.128}\\ \underline{(0.118, 0.139)}\end{tabular} & \begin{tabular}[c]{@{}c@{}}\underline{0.664}\\ \underline{(0.628, 0.701)}\end{tabular} & \begin{tabular}[c]{@{}c@{}}\underline{0.01}\\ \underline{(0.007, 0.013)}\end{tabular}  & \begin{tabular}[c]{@{}c@{}}\underline{0.765}\\ \underline{(0.732, 0.799)}\end{tabular} & \begin{tabular}[c]{@{}c@{}}\underline{0.04} \\ \underline{(0.017, 0.063)}\end{tabular} & \begin{tabular}[c]{@{}c@{}}\underline{0.723}\\ \underline{(0.7, 0.745)}\end{tabular}   & \begin{tabular}[c]{@{}c@{}}\underline{0.019}\\ \underline{(0.013, 0.024)}\end{tabular}  & \begin{tabular}[c]{@{}c@{}}\underline{0.81}\\ \underline{(0.805, 0.815)}\end{tabular}  & \begin{tabular}[c]{@{}c@{}}\underline{0.441}\\ \underline{(0.43, 0.451)}\end{tabular}  & \begin{tabular}[c]{@{}c@{}}\underline{0.666}\\ \underline{(0.652, 0.681)}\end{tabular} & \begin{tabular}[c]{@{}c@{}}\underline{0.636}\\ \underline{(0.613, 0.66)}\end{tabular}  \\ \hline
bioClinicalBERT        & \begin{tabular}[c]{@{}c@{}}0.85\\ (0.84, 0.861)\end{tabular}   & \begin{tabular}[c]{@{}c@{}}0.156\\ (0.138, 0.173)\end{tabular} & \begin{tabular}[c]{@{}c@{}}0.683\\ (0.621, 0.745)\end{tabular} & \begin{tabular}[c]{@{}c@{}}0.008\\ (0.006, 0.011)\end{tabular} & \begin{tabular}[c]{@{}c@{}}0.809\\ (0.785, 0.833)\end{tabular} & \begin{tabular}[c]{@{}c@{}}0.043\\ (0.027, 0.059)\end{tabular} & \begin{tabular}[c]{@{}c@{}}0.76\\ (0.723, 0.796)\end{tabular}  & \begin{tabular}[c]{@{}c@{}}0.02\\ (0.014, 0.027)\end{tabular}   & \begin{tabular}[c]{@{}c@{}}0.83\\ (0.828, 0.831)\end{tabular}  & \begin{tabular}[c]{@{}c@{}}0.469\\ (0.457, 0.48)\end{tabular}  & \begin{tabular}[c]{@{}c@{}}0.68\\ (0.663, 0.697)\end{tabular}  & \begin{tabular}[c]{@{}c@{}}0.653\\ (0.626, 0.68)\end{tabular}  \\
bioGPT                 & \begin{tabular}[c]{@{}c@{}}\textbf{0.862}\\ \textbf{(0.851, 0.872)}\end{tabular} & \begin{tabular}[c]{@{}c@{}}\textbf{0.161}\\ \textbf{(0.141, 0.182)}\end{tabular} & \begin{tabular}[c]{@{}c@{}}0.711\\ (0.679, 0.743)\end{tabular} & \begin{tabular}[c]{@{}c@{}}0.011\\ (0.005, 0.017)\end{tabular} & \begin{tabular}[c]{@{}c@{}}0.818 \\ (0.8, 0.837)\end{tabular}  & \begin{tabular}[c]{@{}c@{}}\textbf{0.047}\\ \textbf{(0.037, 0.058)}\end{tabular} & \begin{tabular}[c]{@{}c@{}}\textbf{0.773}\\ \textbf{(0.734, 0.813)}\end{tabular} & \begin{tabular}[c]{@{}c@{}}\textbf{0.024}\\ \textbf{(0.016, 0.032)}\end{tabular}  & \begin{tabular}[c]{@{}c@{}}\textbf{0.835}\\ \textbf{(0.833, 0.838)}\end{tabular} & \begin{tabular}[c]{@{}c@{}}\textbf{0.478}\\ \textbf{(0.465, 0.492)}\end{tabular} & \begin{tabular}[c]{@{}c@{}}\textbf{0.691}\\ \textbf{(0.672, 0.71)}\end{tabular}  & \begin{tabular}[c]{@{}c@{}}\textbf{0.664}\\ \textbf{(0.638, 0.69)}\end{tabular}  \\
ClinicalBERT           & \begin{tabular}[c]{@{}c@{}}0.855\\ (0.842, 0.867)\end{tabular} & \begin{tabular}[c]{@{}c@{}}0.155\\ (0.137, 0.173)\end{tabular} & \begin{tabular}[c]{@{}c@{}}\textbf{0.717}\\ \textbf{(0.691, 0.743)}\end{tabular} & \begin{tabular}[c]{@{}c@{}}\textbf{0.013}\\ \textbf{(0.009, 0.017)}\end{tabular} & \begin{tabular}[c]{@{}c@{}}0.806\\ (0.784, 0.827)\end{tabular} & \begin{tabular}[c]{@{}c@{}}0.04\\ (0.024, 0.056)\end{tabular}  & \begin{tabular}[c]{@{}c@{}}0.764\\ (0.73, 0.799)\end{tabular}  & \begin{tabular}[c]{@{}c@{}}0.022\\ (0.015, 0.03)\end{tabular}   & \begin{tabular}[c]{@{}c@{}}0.83\\ (0.827, 0.833)\end{tabular}  & \begin{tabular}[c]{@{}c@{}}0.469\\ (0.458, 0.48)\end{tabular}  & \begin{tabular}[c]{@{}c@{}}0.686\\ (0.671, 0.702)\end{tabular} & \begin{tabular}[c]{@{}c@{}}0.66\\ (0.634, 0.686)\end{tabular}  \\ \bottomrule
\end{tabular}}
\caption{. A comparison of traditional NLP models (top) vs pretrained LLMs (bottom). The results are presented as the mean and 95\% confidence interval across all 5-folds. The best baseline models are \underline{underlined}, and the best models are \textbf{bolded}. As shown amongst the results, the pretrained LLMs consistently outperform traditional word embeddings.} \label{table:embedding_vs_pretrained}
\end{sidewaystable}

When comparing the predictive performances of clinically-oriented pretrained LLMs, such as bioGPT \cite{luo2022biogpt}, ClinicalBERT \cite{huang2019clinicalbert}, and bioClinicalBERT \cite{alsentzer2019publicly}, with traditional word embeddings, including word2vec's continuous bag-of-words (CBOW) \cite{Church_word2vec_2016}, doc2vec \cite{Wijffels_dov2vec_2020}, GloVe \cite{Pennington_glove_2014}, and FastText \cite{Athiwaratkun_fasttext_2018}, we observed absolute increases that ranged from up to 20.7\% in delirium to 38.3\% in death within 30 days for the Area Under the Receiver Operating Characteristic curve (AUROC). Similarly, increases in the Area Under the Precision-Recall Curve (AUPRC) ranged from 0.9\% in PE to an impressive 33.2\%.

This massive leap in performance highlights the sheer power of pretrained LLMs in grasping clinically relevant context in comparison to traditional word embeddings, despite being trained on broad biomedical and/or clinical notes that are not specific to perioperative care. These improvements, experimented akin to a `zero-shot setting' \cite{brown_fewshot_2020}, are noteworthy given their minimal exposure to perioperative care notes.

\subsection{Transfer learning: Improvements from adapting pretrained models to perioperative corpora}
\label{label:self_supervised}
\begin{figure}
\hspace*{-3.5cm}
    \includegraphics[width=1.55\textwidth,height=\textheight,keepaspectratio]{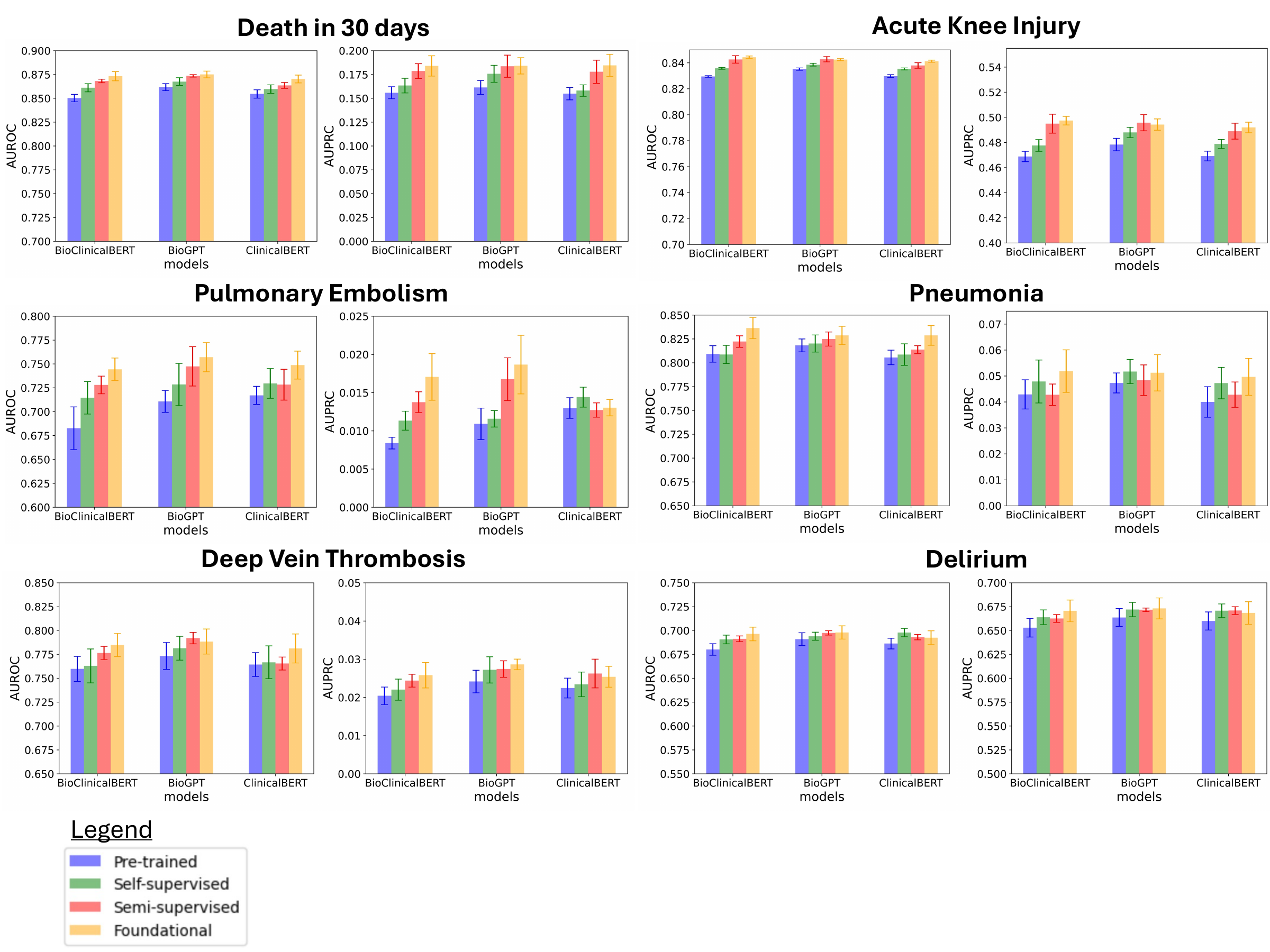}
    \caption{Comparison of the predictive performance across various models and their respective tuning strategies. The bar graph illustrates the means, with the error bar representing the respective standard errors, across a 5-fold cross-validation. Fine-tuning with the models self-supervised pretraining objectives improved prediction performance relative to the pretrained models alone, with the incorporation of labels further boosting prediction performances. The model performs best with the foundation fine-tuning strategy, wherein the model was fine-tuned with a multi-task learning objective across all outcomes. Precise numerical metrics are reported in the supplementary material.}
    \label{fig:results}
\end{figure}

\label{sec:self_supervised}

Exposing each of the selected clinically oriented pretrained models -- bioGPT, ClinicalBERT, and bioClinicalBERT -- to clinical texts specific to perioperative care through self-supervised fine-tuning, as best illustrated in Figure \ref{fig:finetuning_methods}, resulted in absolute improvements ranging from up to 0.6\% in AKI to 3.2\% in PE for AUROC compared to using the pretrained models demonstrated in Section \ref{sec:pretrained_vs_word}. Similarly, increases ranged from up to 0.3\% in PE to 1.5\% in 30-day mortality for AUPRC. The results spanning the three base models -- bioClinicalBERT, bioGPT and ClinicalBERT -- across all outcomes are best illustrated in Figure \ref{fig:results}. These consistent improvements witnessed upon adapting these pretrained models to perioperative specific corpora highlight the gap between the corpora on which the pretrained models were originally trained and that of perioperative care. This suggests that there remains leeway for improvement in adapting the models to the textual characteristics specifically observed in perioperative care notes.

\subsection{Transfer learning: Improvements from further incorporating labels}


When incorporating labels as part of the fine-tuning process through a semi-supervised approach, we witnessed further improvements ranging from up to 0.4\% in delirium to 1.8\% in PE, and AUPRC values ranging from 0\% in Delirium to 2\% in death within 30 days, relative to the approach in section \ref{sec:self_supervised}. The results spanning the three base models across all outcomes is best illustrated in Figure \ref{fig:results}. This demonstrates that predictive performance is boosted relative to the self-supervised approach since both textual and labelled data were utilized during training. 

\subsection{The foundational capabilities of LLMs in predicting post-operative complications from pre-operative notes}
\label{sec:foundational}

Foundation fine-tuning involves incorporating a multi-task learning framework to simultaneously fine-tune all labels, resulting in a single robust model capable of predicting all six possible postoperative risks. Through our foundation model, we observe increases in AUROC ranging from up to 0.8\% in AKI to 3.6\% in PE and AUPRC values increases ranging from up to 0.4\% in Pneumonia to 2.6\% in death within 30 days, when contrasted with the self-supervised approach from section \ref{label:self_supervised}. The results spanning the three base models across all outcomes is best illustrated in Figure \ref{fig:results}. As this is our best-performing finetuning strategy, we reported additional metrics, such as the accuracy, sensitivity, specificity, precision, and F-scores, in the supplementary material.

The attainment of competitive performances are a testament towards the ability of LLMs to establish robust representations across varied clinical tasks, therefore showcasing the foundational capabilities of LLMs in perioperative care.

\subsection{Examining the effects of ML predictors on predictive performance}

\begin{figure}
\hspace*{-3.5cm}
    \includegraphics[width=1.55\textwidth,height=\textheight,keepaspectratio]{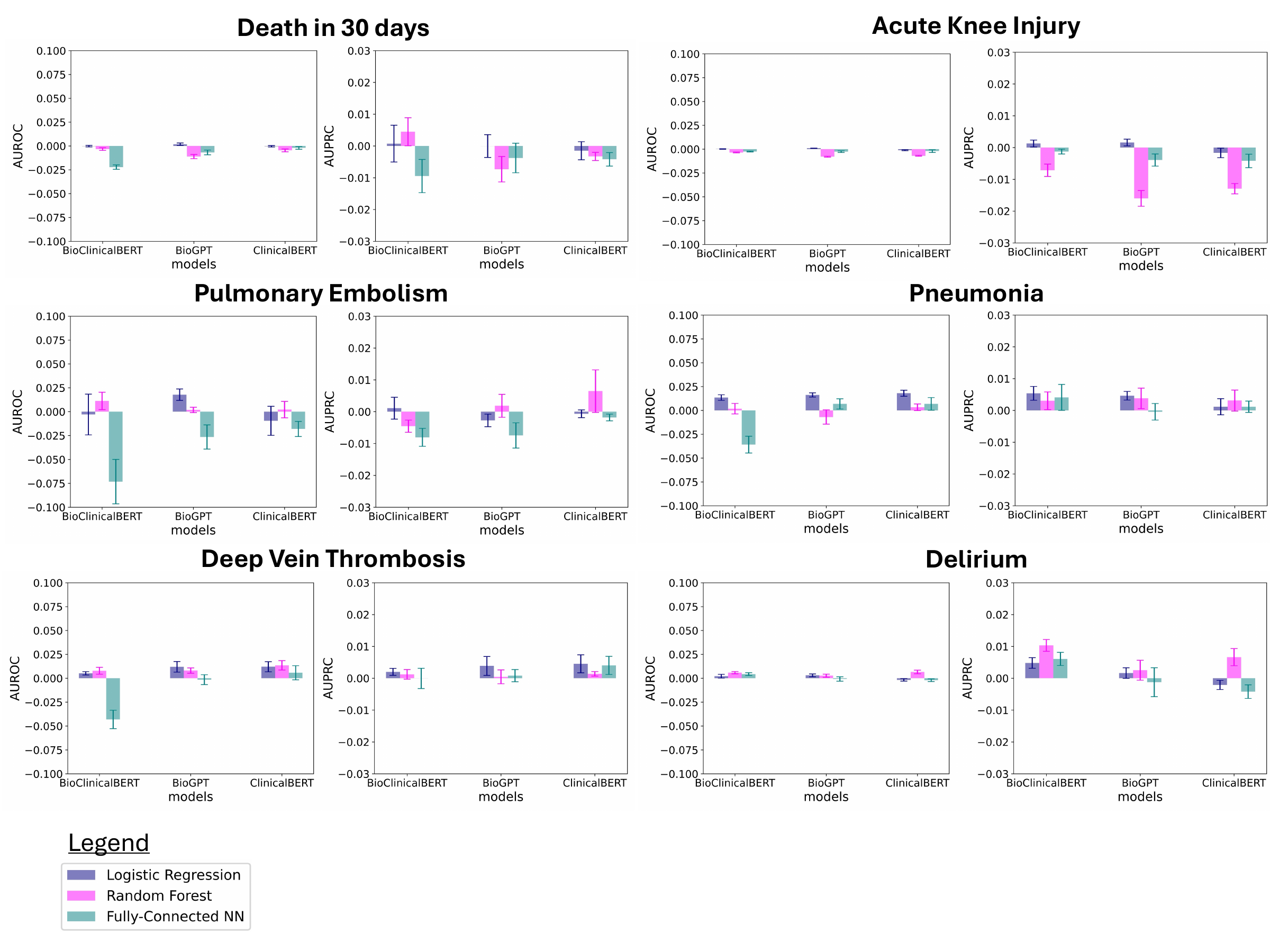}
    \caption{Comparison of different machine learning classifiers with that of our default XGBoost predictor applied to our textual representations ($\Delta \text{model}_{i,j} - \text{XGBoost}_i$), including the use of the feed-forward auxiliary layer directly from our foundation model. No single classifier dominated the others across all outcomes and metrics. Surprisingly, the logistic regression classifier performed slightly better than the others, demonstrating that well-tuned language models can generate precise contextual representations to suit a simple classifier. Precise numerical metrics are reported in the supplementary material.}
    \label{fig:predictors}
\end{figure}

To facilitate consistent comparisons between the traditional word embeddings and the distinct fine-tuning strategies from pretrained LLMs, we defaulted to using eXtreme Gradient Boosting Tree (XGBoost) \cite{Chen_XGBoost_2016} as the predictor used for the text embeddings. However, in practice, the choice of predictors remains flexible. 

As such, we compare the performances of applying distinct ML predictors among the best-performing fine-tuning strategy -- foundation models. These include (1) the default XGBoost, (2) logistic regression, (3) random forest, and (4) task-specific fully connected neural networks found on top of the contextual embeddings. The results are mixed, with no single classifier dominantly outperforming the others. Surprisingly, logistic regression performed slightly better than the others, demonstrating that well-tuned models can generate precise contextual representations that suit a simple classifier with better generalizability than complex classifiers.

\subsection{Qualitative evaluation of our foundation models for safety and adaption to perioperative care}

We ran several broad case prompts to examine and ensure the safety \cite{Farquhar_2024,Van_2024} of our foundation models. This also provides us with the opportunity to qualitatively ensure its adaptation to perioperative care beyond the reported quantitative results.

\begin{table}[]
\textbf{(a) BioClinicalBERT variant}
\resizebox{\textwidth}{!}{
\begin{tabular}{lll}
\toprule
Prompt                                                                                                                            & bioClinicalBERT                                                                                                               & Our   model                                                                                                                     \\ \midrule \\
"{[}MASK{]} underwent surgery to remove tumor."                                                                                   & \underline{\textbf{She}} underwent surgery to remove tumor.                                                                                  &  \underline{\textbf{Patient}} underwent surgery to remove tumor.                                                                           \\~\\
\begin{tabular}[c]{@{}l@{}}``Patient complains about pain. \\ Patient is {[}MASK{]}."\end{tabular}                                 & \begin{tabular}[c]{@{}l@{}}Patient complains about pain. \\ Patient is \color{red}{\underline{\textbf{comfortable}}}.\end{tabular}                              & \begin{tabular}[c]{@{}l@{}}Patient complains about pain. \\ Patient is \underline{\textbf{anxious}}.\end{tabular}                                    \\~\\
\begin{tabular}[c]{@{}l@{}}``Patient suffers from excessive {[}MASK{]}. \\ Patient should be   sent to ICU."\end{tabular}          & \begin{tabular}[c]{@{}l@{}}Patient suffers from excessive \underline{\textbf{anxiety}}. \\ Patient should be sent to ICU.\end{tabular}             & \begin{tabular}[c]{@{}l@{}}Patient suffers from excessive \underline{\textbf{burn}}. \\ Patient should be sent to ICU.\end{tabular}                   \\~\\
\begin{tabular}[c]{@{}l@{}}``Patient's family has history of diabetes. \\ Surgeon should   watch out for {[}MASK{]}."\end{tabular} & \begin{tabular}[c]{@{}l@{}}Patient's family has history of diabetes. \\ Surgeon should watch   out for \underline{\textbf{diabetes}}.\end{tabular} & \begin{tabular}[c]{@{}l@{}}Patient's family has history of diabetes. \\ Surgeon should watch   out for \underline{\textbf{procedures}}.\end{tabular} \\~\\ \bottomrule
\end{tabular}}
\\~\\~\\
\textbf{(b) BioGPT variant}
\resizebox{\textwidth}{!}{
\begin{tabular}{lll}
\toprule
Prompt                                                                                                    & bioGPT                                                                                                                                                      & Our   model                                                                                                                                              \\ \midrule
\begin{tabular}[c]{@{}l@{}}Patient suffers from excessive bleeding. \\ Patient should be \dots~\dots\end{tabular} & \begin{tabular}[c]{@{}l@{}}Patient suffers from excessive bleeding. \\ Patient should be \underline{\textbf{sent to a hospital}} \\ \underline{\textbf{for further investigation.}}\end{tabular}      & \begin{tabular}[c]{@{}l@{}}Patient suffers from excessive bleeding. \\ Patient should be \underline{\textbf{sent to a trauma room }}\\ \underline{\textbf{or a vascular access room}} \end{tabular} \\~\\
\begin{tabular}[c]{@{}l@{}}Patient complains of extreme pain. \\ Patient is probably a \dots~\dots \end{tabular}   & \begin{tabular}[c]{@{}l@{}}Patient complains of extreme pain. \\ Patient is probably a \textcolor{red}{\underline{\textbf{good candidate for a }}}\\ \textcolor{red}{\underline{\textbf{trial of conservative treatment}}}\end{tabular} & \begin{tabular}[c]{@{}l@{}}Patient complains of extreme pain. \\ Patient is probably a case of \underline{\textbf{lumbar}}\end{tabular}                                       \\~\\
\begin{tabular}[c]{@{}l@{}}Patient's family has history of \\ high blood pressure. Avoid \dots~\dots\end{tabular} & \begin{tabular}[c]{@{}l@{}}Patient's family has history of \\ high blood pressure. Avoid \\ \underline{\textbf{use of antihypertensive drugs}}.\end{tabular}                     & \begin{tabular}[c]{@{}l@{}}Patient's family has history of \\ high blood pressure. Avoid \\ \underline{\textbf{invasive diagnostic procedure}} \\ \underline{\textbf{if possible}}\end{tabular} \\ \bottomrule  
\end{tabular}}
\caption{A qualitative safety evaluation towards the open-sourced (a) bioClinicalBERT and (b) bioGPT variant of our foundation models. The prompts demonstrate the safety and adaptive nature of our model in perioperative care use. Results deemed potentially harmful are colored in \color{red}{red}.}

\label{table:qualitative_eval}
\end{table}

Table \ref{table:qualitative_eval} demonstrates that our foundation models have indeed adapted from the generic texts in biomedical literature or non-perioperative clinical notes to the terminologies expected among perioperative care notes. More importantly, our model did not produce any alarmingly harmful outputs, contrary to what we witnessed in the outputs among some of the pretrained models of which our foundation models were based on. 
\subsection{BERT vs GPT / biomedical vs clinical-domain based models -- Which performs best in predicting postoperative risks?}

From the pretrained models alone (i.e., without any fine-tuning), BioGPT excelled in 5 of the 6 tasks, while ClinicalBERT excelled in 1 of the 6 tasks. This demonstrates that the text and terminologies in biomedical literature sufficiently capture the text and terminologies of a similar magnitude compared to their counterparts trained on critical care unit notes.

However, upon fine-tuning the models with the clinical notes and the six labels, the foundation BioGPT model excelled in 4 of the 6 tasks, while BioClinicalBERT excelled in 2 of the 6 tasks. This can be attributed to the fact that the training texts from the biomedical corpora from the base model, was complemented by terminologies and texts infused from our perioperative care notes during fine-tuning, thereby creating robust, diverse, and rich representations compared to notes solely exposed to clinical corpora as in the case of the ClinicalBERT version of our foundation models.

Minimal differences were also observed between the encoder-only Bidirectional Encoder Representations from Transformers (BERT) models \cite{BERT_2019} and the decoder-only Generative pretrained Transformer (GPT) models \cite{GPT_2019}, in-spite the BERT models being more tailored to downstream classification tasks in other domains, such as sentiment classification and name entity recognition (NER) \cite{10092289}. This suggests that GPT models, traditionally used for generative tasks, can also perform equally or more competitively in classification tasks in perioperative care contrary to what is expected of other domains.

\subsection{Generalizability of our methods}

Generalizability of methods across datasets spanning distinct health centers and hospital units remains a core concern in building predictive models towards clinical care \cite{Goetz_2024}. To demonstrate that our methods are generalizable beyond BJH and perioperative care, we replicate our methods towards the publicly available MIMIC-III \cite{Johnson_MIMIC_2016} dataset. Details of the results and the specifics aspects of MIMIC-III that were used to replicate our methods are detailed in the supplementary material. Our results aligned with the main findings of our study. Specifically, the baseline models were outperformed by pretrained LLMs, with absolute increases that ranged from up to 14.6\% in 12-hour discharge to 30\% for In-hospital mortality for AUROC. Similarly, increases in the AUPRC ranged from 17.7\% in 12-hour discharge to 28.2\% for in-hospital mortality. Similar to the results observed in our BJH dataset, self-supervised fine-tuning witness maximal absolute improvements in AUROCs of up to 0.4\% in 12-hour discharge to 3\% in in-hospital mortality and AUPRCs of up to 0.4\% in 30-day mortality to 0.8\% in 12-hour discharge. Incoporating labels into the fine-tuning process also saw further improvements of 0.5\% in 12-hour discharge to 1.6\% in 30-day mortality and 0.3\% in 12-hour discharge to 3.6\% for in-hospital mortality for AUROC and AUPRC, respectively. Finally, foundation models again performed the best, with AUROC improvements of 0.4\% in 12-hour discharge to 1.4\% in 30-day mortality and AUPRC improvements of 0.2\% in 12-hour discharge to 2.6\% for in-hospital mortality when compared to self-supervised fine-tuning. In the same order, the mean squared error (MSE) of LOS decreased by up to 85.9 days, 3.1 days, 8 days and 8.2 days, respectively.

The replicability of methods demonstrates that the foundational capabilities of LLMs potentially extends beyond perioperative care and BJH to other domains of clinical care spanning different healthcare systems.

\section{Discussion}\label{sec3}

The release of ChatGPT and its rapid rise in popularity have sparked significant interest in applying LLMs to decision support systems in medical care \cite{Raza_Venkatesh_Kvedar_2024, Wornow_2023}. However, challenges such as the low-resource nature of clinical notes -- which are neither readily open-sourced nor abundantly available \cite{wu2022survey} -- and the complex abbreviations and specialized medical terminologies that are not well-represented in the training corpora of pre-trained models \cite{Wornow_2023} have made the applications of LLMs in healthcare onerous. These obstacles have driven efforts to refine, adapt, and evaluate LLMs for reliable use across distinct clinical environments, such as personalized treatment in patient care \cite{yang2023exploring}, the automation of medical reports \cite{wang2023chatcad} or summaries \cite{Van_2024}, and medical diagnosis for conditions like COVID-19 \cite{li2023text} or Alzheimer's \cite{mao2023ad}. Despite these advancements, the application of LLMs in forecasting postoperative surgical complications from preoperative notes remains largely unexplored. Addressing this gap is critical, as such predictions could facilitate the early identification of patient risk factors and enable the timely administration of evidence-based treatments, such as antibiotics in postoperative settings \cite{Xue_2021b, Bland_2006, Young_Khadaroo_2014}.

While various AI models leveraging discrete EHR data have demonstrated success in the early identification of post-operative risks \cite{Xue_2021b,Xue_perioperative_2022}, clinical notes taken during the preoperative phase hold immense informational value that is often not captured through tabular data alone. This underscores the pressing need to develop decision support systems focused on text-based preoperative notes. To address this need, our retrospective study utilizes open-source, pre-trained clinical LLMs on pre-operative clinical notes to predict postoperative surgical complication. Drawing on approximately 85,000 notes from Barnes-Jewish Hospital (BJH) -- the largest hospital in Missouri and the largest private employer in the Greater St. Louis region \cite{BJH_hospital} -- our study represents a crucial step forward in integrating LLMs into perioperative decision-making processes. By supporting surgical care management, optimizing perioperative care, and enhancing early risk detection, LLMs have the potential to improve patient outcomes and reduce complications that could be avoided through early identification of patient risk factors.

Advanced improvements observed when using pretrained clinical LLMs compared to traditional word embeddings underscore the capability of such models without needing model training, eliminating the potential constraints in training data collection and computational costs and underscoring their capability adjacent to a `zero-shot' setting \cite{brown_fewshot_2020}. These results are noteworthy given differences between the training copora of such models, such as PubMed articles or public EHR databases like MIMIC, and the semantic characteristics of our clinical notes.  Such findings may be handy when data or computational resources may be limited \cite{sandmann2024systematic}. Additionally, the enhancements through transfer learning via fine-tuning in an self-supervised setting could be attributed to the adaption of these pretrained models in `familiarizing' itself towards the texts of our preoperative clinical notes \cite{BERT_2019,GPT_2019}. Such differences might arise from variations in text distributions or frequency, as well as the distinctive use of abbreviations and terminology in each task-specific dataset, contrasting the textual contents found in databases of pretrained models like PubMed or Wikipedia \cite{Wornow_2023,Van_2024, shojaee2024evaluating}. This perspective highlights the potential limitations of relying purely entirely on pretrained models to capture precise context within diverse clinical texts and accentuates the necessity to refine these LLMs using more extensive clinical datasets \cite{Wornow_2023,shojaee2024evaluating}. This thereby shows that even in cases where labelled data is scarce, tuning the models with respect to the clinical texts can still lead to further improvements compared to using pretrained models alone. Incorporating labels into the fine-tuning process, when available, can substantially improve performance relative to self-supervised fine-tuning. This ensures tokens linked to specific outcomes are effectively represented in the optimized model, thereby achieving improved performances \cite{huang2019clinicalbert,chen_fine_2020}. The foundation models reached peak performance by incorporating a multi-task learning (MTL) framework at scale, as concurred by Aghajanyan et al \cite{Aghajanyan_2021}. Henceforth, a model-agnostic foundation language model – where a single robust model is fine-tuned across all accessible labels -- yielded more competitive outcomes. Such performance is a testament to the model's ability to efficiently capture and retain crucial predictive information through knowledge sharing among distinct postoperative complications during the fine-tuning phase, thereby establishing the foundational capabilities of LLMs in predicting risks from notes taken during perioperative care. 

The foundation approach offers tangible benefits beyond achieving strong results. From a computational standpoint, a single model can be fine-tuned to serve various tasks, conserving time and resources that would be expended on tailoring and storing parameters for each specific model \cite{Aghajanyan_2021,Raza_Venkatesh_Kvedar_2024, perez2024guide}. Additionally, this model can generalize across tasks it has been optimized for through knowledge sharing, outperforming bespoke models and mitigating concerns of potential overfitting to the limited samples from a specific task \cite{kernbach2022foundations}. From a clinical standpoint, because a single robust model can be deployed to predict multiple tasks, it can be conveniently integrated into clinical workflows \cite{Raza_Venkatesh_Kvedar_2024, perez2024guide}. This versatility not only enhances the utility of the model but also contributes to a reduction in the cognitive load required by clinicians to decipher potential risk factors manually \cite{koopman2015physician}, thereby reducing the risk of clinician burnout, which is presently at near crisis levels \cite{lou2022predicting}. This simultaneously mitigates issues surrounding the lack of established routines for reading clinical notes, the lack of knowledge in the existence of notes taken prior to handoffs, and competing time demands, which are key individual factors leading to unintentional missed recommendations \cite{Flemons_2022}. Clinicians could henceforth shift their focus to designing and implementing interventions to mitigate these risks, which could be vital in reducing post-operative complications \cite{centre2020impact,Wolfhagen_2022}. This is supported by studies have shown that carefully designed interventions from the early identification of risk factors could mitigate complications by up to 80\%, which is more effective than the effect of drug or treatment interventions \cite{centre2020impact}.

Minor variances in outcomes emerged when comparing the results of training different machine learning predictors on extracted textual embeddings, including the utilization of the fully connected neural network found atop the contextual representations. This affirms that the robustness of LLMs lies in the attention-based architecture, which leads to precise text representations, while the domain-specific dataset on which the language model is fine-tuned plays a crucial role in its adaptability \cite{Johnson_MIMIC_2016,lee2020biobert,alsentzer2019publicly}. Similarly, there was no consistent performance gain using BERT-based clinical LLM models over GPT-based models, even though they had different training strategies and the former is more widely used in downstream prediction. This highlights the need for ongoing enhancement of clinical LLMs using innovative techniques across a wide spectrum of medically relevant datasets \cite{Wornow_2023}, rather than focusing on the predictor model’s training architecture.

Given that model harmfulness and hallucinations are core concerns in the application of LLMs in medicine \cite{Van_2024}, we tested our model with several broad-case prompts. Our outputs not only assured us of its safety, contradicting some of the potentially harmful outputs witnessed from the base models our foundation models were trained on, but also qualitatively assert its semantic shift towards perioperative-specific corpora from the broad biomedical or critical care unit notes on which these base models were trained \cite{Wornow_2023}.

Despite demonstrating the potential of LLMs in predicting risks from notes taken during preoperative care, our study possesses limitations. First, quality of the textual data may possess limitations in itself, ranging from incompleteness \cite{troiani2005incomplete}, to potentially missing data \cite{troiani2005incomplete}, to the potential lack of frequent updates \cite{walker2019opennotes}, which could potentially possess an adverse impact on the performances of our fine-tuned LLMs. Second, non-textual variables were not accounted for in the models, including demographics, preoperative measurements, key intraoperative variables (e.g., blood transfusion data, and urine output), and certain medications used during surgery \cite{Xue_2021b}. Adding these variables could potentially improve the model performance \cite{Xue_2021b}. Third, it remains unclear if the observations can be generalized to LLMs in other applications or at various scales (e.g., the public LLaMa models with various sizes). Forth, subgroup analysis based on the various surgery types was not conducted because of the small number of patients within each subgroup; hence, the clinical utility of the predictions of postoperative complications based on specific surgery types is limited. To address these limitations, our ongoing work involves data triangulation across the administrative data, clinical text, and other data to align with high-quality manual health record review provided by National Surgical Quality Improvement Program adjudicators.

\section{Methods}

\subsection{Data}
Our dataset originated from electronic anesthesia records (Epic) for all adult patients undergoing surgery at Barnes-Jewish Hospital (BJH), the largest healthcare system in the greater St. Louis (MO) region, spanning four years from 2018 to 2021. The dataset included 
$n=84,875$ surgical cases. The text-based notes were single-sentence documents with a vocabulary size of $|V|=3,203$, and mean word and vocabulary lengths of $\overline{l}_w = 8.9$ (SD: 6.9) and $\overline{l}_v = 7.3$ (SD: 4.4), respectively. The textual data contained detailed information on planned surgical procedures, which were derived from smart text records during patient consultations. To preserve patient privacy, the data was provided in a de-identified and censored format, with patient identifiers removed and procedural information presented in an arbitrary order, ensuring that no information could be traced back to any uniquely identifiable patient.

\subsection{Participants}
Of the 84,875 patients, the distribution of patient types included 17\% (14,412) in orthopedics, 8.8\% (7,442) in ophthalmology, and 7.4\% (6,236) in urology. The gender distribution was 50.3\% male (42,722 patients). Ethnic representation included 74\% White (62,563 patients), 22.6\% African American (19,239 patients), 1.7\% Hispanic (1,488 patients), and 1.2\% Asian (1,015 patients). The mean weight was 86 kg ($\pm$24.7 kg) and the mean height was 170 cm ($\pm$11 cm). Other characteristics of the patients could be found in Table \ref{table:characteristics}. The internal review board of Washington University School of Medicine in St. Louis (IRB \#201903026) approved the study with a waiver of patient consent.

\begin{table}[]
\centering
\begin{tabular}{cc}
\toprule
Characteristics & Statistic \\
\midrule
                                                     
\multirow{3}{*}{Patient type}    & Orthopedic 14412 (17\%)              \\
                                 & Ophthalmology:   7442 (8.8\%)            \\
                                 & Urology:   6236 (7.4\%)           \\~\\
Gender                           & Male: 42722 (50.3\%)               \\~\\
\multirow{4}{*}{Ethnicity}       & White: 62563 (74\%)                    \\
                                 & African   American: 19239 (22.6\%),  \\
                                 & Hispanic:   1488 (1.7\%)           \\
                                 & Asian:   1015 (1.2\%)                  \\~\\
Weight                           & 86kg (24.7kg)                      \\~\\
Height                           & 170 cm (11cm)                         \\~\\
Liver disease                    & Yes: 6697 (7.9\%)                     \\~\\
Cancer                           & Yes: 29213 (34\%)                     \\~\\
Congestive Heart Failure         & Yes: 8886 (10\%)                  \\~\\
Myocardial Infarction            & Yes: 8587 (10\%)                    \\~\\
Chronic Pulmonary Disease        & Yes: 9837 (12\%)                       \\~\\
HIV/AIDS                         & Yes: 4069 (4.8\%)                      \\ \bottomrule
\end{tabular}
\caption{Characteristics of patients from the BJH dataset. Categorical variables are reported as number of patients (percentage), numerical variables are reported as mean (standard deviation). Note that the summary statistics were computed after removing records with no texts associated with the patient.}
\label{table:characteristics}
\end{table}
\clearpage

\subsection{Outcomes}

Our six outcomes were: 30-day mortality, acute knee injury (AKI), pulmonary embolism (PE), pneumonia, deep vein thrombosis (DVT), and delirium. These six outcomes remain pertinent in perioperative care, particularly during OR-ICU handoff \cite{Bellini_2022,Xue_2021b, amin2019reducing}.

AKI was determined using a combination of laboratory values (serum creatinine) and dialysis event records, and structured anesthesia assessments, laboratory data, and billing data indicating baseline end-stage renal disease were used as exclusion criteria for AKI. Acute kidney injury was defined according to the Kidney Disease Improving Global Outcomes criteria. Delirium was determined from nurse flow-sheets (positive Confusion Assessment Method for the Intensive Care unit test result); pneumonia, DVT, and PE were determined based on the International Statistical Classification of Diseases and Related Health Problems, Tenth Revision (ICD-10) diagnosis codes. Patients without delirium screenings were excluded from the analysis of that complication.

\subsection{Models}

\subsubsection{Traditional word embeddings and pretrained LLMs}
\label{sec:traditional_and_pretrained_models}
We employed a combination of BERT \cite{BERT_2019} and GPT-based \cite{GPT_2019} large language models (LLMs) trained on either biomedical or open-source clinical corpora --specifically BioGPT \cite{luo2022biogpt}, ClinicalBERT \cite{huang2019clinicalbert}, and BioClinicalBERT \cite{alsentzer2019publicly} -- for predicting risk factors from notes taken during perioperative care. BioGPT is a 347-million-parameter model trained on 15 million PubMed abstracts \cite{luo2022biogpt}. It adopts the GPT-2 model architecture, making it an auto-regressive model trained on the language modeling task, which seeks to predict the next word given all preceding words. In contrast, ClinicalBERT was trained on the publicly available Medical Information Mart for Intensive Care III (MIMIC-III) dataset, which contains 2,083,180 de-identified clinical notes associated with critical care unit admissions \cite{huang2019clinicalbert}. It was initialized from the $\text{BERT}_{\text{base}}$ architecture with the masked language modeling (MLM) and next sentence prediction (NSP) objectives \cite{BERT_2019}. Similarly, BioClinicalBERT was pretrained on all the available clinical notes associated with the MIMIC-III dataset \cite{alsentzer2019publicly}. However, unlike ClinicalBERT, BioClinicalBERT was based on the BioBERT model \cite{lee2020biobert}, which itself was trained on 4.5 billion words of PubMed abstracts and 13.5 billion words of PubMed Central full-text articles. This allowed BioClinicalBERT to leverage texts from both the biomedical and clinical domains.

These models have been tested across representative NLP benchmarks in the medical domains, including Question-Answering tasks benchmarked by PubMedQA \cite{luo2022biogpt,jin_2019_pubmedqa}, recognizing entities from texts \cite{alsentzer2019publicly} and logical relationships in clinical text pairs \cite{shivade2019mednli}. In addition to the clinically-oriented LLMs, we included traditional NLP word embeddings as baselines for comparison. These include word2vec's continuous bag-of-words (CBOW) \cite{Church_word2vec_2016}, doc2vec \cite{Wijffels_dov2vec_2020}, GloVe \cite{Pennington_glove_2014}, and FastText \cite{Athiwaratkun_fasttext_2018}

By comparing these transformer-based LLMs with traditional words embeddings, we seek to establish the magnitude of improvements clinically-oriented pretrained LLMs could potentially possess in grasping and contextualizing perioperative texts towards predicting postoperative risks in comparison to traditional word embeddings, represents each word as a vector treated as an independent entity.

\subsubsection{Fine-tuned models}

A comparison of the distinct fine-tuning methods employed among our study could be best illustrated in Figure \ref{fig:finetuning_methods}. 

\paragraph{Transfer learning: self-supervised fine-tuning}

To bridge the gap between the corpora of the pretrained models and that of perioperative notes, we first expose and adjust these models to our perioperative text through self-supervised fine-tuning. We adapt the pretrained LLMs with our training data in accordance with their existing training objectives. This process leverages the information contained within the source domain and exploits it to align the distributions of source and target data. For BioGPT, this entails the language modeling task \cite{luo2022biogpt, GPT_2019}. For ClinicalBERT and BioClinicalBERT, it involves the masked language modeling (MLM), as well as the Next Sentence Prediction (NSP) objectives if the document contains multiple sentences, as elaborated in section \ref{sec:traditional_and_pretrained_models} \cite{BERT_2019}. The fine-tuning parameters, including number of epochs, batch size, weight decay, and learning rate, for each of the three base models, are detailed in the supplementary material.

\paragraph{Transfer learning: incorporating labels into fine-tuning}
In lieu of anticipated improvements from self-supervised fine-tuning, we took it a step further by incorporating labels as part of the fine-tuning process, in the hopes of boosting performance as demonstrated in past studies \cite{chen_fine_2020,huang2019clinicalbert}. We do this through a semi-supervised approach, as illustrated in Figure \ref{fig:finetuning_methods}. This means that in contrast to the self-supervised approach which adjusts weights based solely on training texts, the semi-supervised method infuses label information during the fine-tuning process. In doing so, the model leverages an auxillary fully-connected feed-forward neural network atop of the contextual representations, found in the final layer of the hidden states, to predict the labels as part of its fine-tuning process. The auxillary neural network uses the Binary-Cross-Entropy (BCE), Cross-Entropy (CE), and Mean-Square-Error (MSE) losses for binary classification, multi-label classification, and regression tasks, respectively. A $\lambda$ parameter was introduced to balance between magnitude of losses between the supervised and self-supervised objectives. The $\lambda$ parameter, as well as all other parameters used to fine-tune the models, are detailed in the supplementary material. Henceforth, in addition to the potential improvements brought by the self-supervised training objectives, we are now able to leverage the labels to supervise the textual embedding to better align with the training labels. 

\paragraph{foundation fine-tuning strategy}

To build a foundation model with knowledge across all tasks, we extended the above-mentioned strategies and exploited all possible labels available within the dataset, including but not limited to selected tasks, as inspired by Aghajanyan et al \cite{Aghajanyan_2021}. This involves employing a multi-task learning framework for knowledge sharing across all available labels from all six outcomes -- death in 30 days, AKI, PE, pneumonia, DVT, and delirium -- present in the dataset. This task-agnostic approach therefore enables knowledge sharing across all available labels. Therefore, the model becomes foundational in the sense that it solves various tasks simultaneously, meaning a single robust model can be deployed to a wide range of postoperative risks. This contrasts with previous approaches that required a separate model dedicated to each specific outcome. To achieve this, each label is assigned a task-specific auxiliary fully-connected feed-forward neural network, wherein the losses across all labels are pooled together. To control for the magnitude of losses between each task-specific auxiliary network, $\lambda_i$ for $i=1, \dots , m$ given $m$ outcomes, is introduced as weights that contribute to the overall loss calculation.  $\lambda_i$, as well as all other parameters used to fine-tune the models, are detailed in the supplementary material.

\subsection{Predictors}

We defaulted to using XGBoost \cite{Chen_XGBoost_2016} to predict the outcomes from the generated word embeddings or text representations to facilitate consistent comparisons between traditional word embeddings, pretrained LLMs, and their fine-tuned variants. This selection allows to accommodate a diverse range of input types while leveraging XGBoost's widespread use in healthcare due to its robust performance in various clinical prediction tasks. Nonetheless, the choice of predictors remain flexible. This includes utilizing the task specific fully-connected feed-forward neural network found in models that were fine-tuned with respect to the labels. 

To examine if the choice of predictors makes a noticable difference in predictive performances, we experimented with various predictors among the models of the best performing fine-tuning strategy. These include (1) the default XGBoost, (2) Random Forest, (3) Logistic Regression (4) and the feed-forward network found in models that incorporate labels into the fine-tuning process. The range of hyper-parameters used for each predictor could be found in the supplementary material.

\subsection{Evaluation metrics and validation strategies}

For a rigorous evaluation, experiments were stratified into 5 folds for cross-validation \cite{kernbach2022foundations}. A nested cross-validation approach was used to ensure a robust and fair comparison of the best combination of hyperparameters across all models and approaches spanning multiple folds. 

The main evaluation metrics included area under the receiver operating characteristic curve (AUROC) and the area under the precision-recall curve (AUPRC) to get a comprehensive evaluation of the models' overall prediction performance in the face of class imbalance. For our best-performing foundation models, we also reported their accuracy, sensitivity, specificity, precision, and F-scores, which could be found in the supplementary material.

\subsection{Qualitative evaluation of our foundation models}

We qualitatively evaluate our models for safety while ensuring their adaptation to perioperative corpora beyond the quantitative results, prior to open-sourcing them. As such, we ran some broad case-prompts on our best-performing foundation models through their objective loss functions. For the BERT-based models, this encompassed the masked language modeling (MLM) objective, where we get the model to fill a single masked token in a `fill-in-the-blank' format \cite{BERT_2019}. For the GPT models, this involved the language modeling objective, where we get the model to `complete the sentence' based on the provided incomplete sentences \cite{GPT_2019}.

\subsection{Replication of methods on MIMIC-III}

To ensure that our methods are generalizable beyond perioperative care and the BJH dataset, we replicated our methods on the publicly available MIMIC-III dataset \cite{Johnson_MIMIC_2016}, which contains de-identified reports and clinical notes of critical care patients at the Beth Israel Deaconess Medical Center between 2001 and 2012. To closely mimic the approach and settings employed in BJH's clinical notes, we utilized the long-form descriptive texts of procedural codes in MIMIC-III. Specifically, ICD-10 codes containing procedural information from each patient were traced and formatted into their respective long-form titles. For each patient, these long-form titles were then combined to form a single-sentenced clinical note, thereby aligning with the textual characteristics of BJH's clinical notes. The selected outcomes were length-of-stay (LOS), in-hospital mortality, 12 hour discharge status, and death in 30 days, as adapted from previous studies \cite{Johnson_MIMIC_2016,luo2022association, Xue_2021b}. Details on the data characteristics and extraction of outcomes could be found in the supplementary material.

\bmhead{Supplementary Information}

Supplementary material have been attached and provided alongside the manuscript submission to the journal. 

\bmhead{Acknowledgements}

The authors would like to thank Brad Fritz, Michael Avidan, Christopher King, and Sandhya Tripathi from Washington University for collecting and providing the dataset.

\section*{Data availability}
Our data could not be shared per our IRB. A Patient waiver was obtained for the study. Nonetheless, the best performing foundation models are gated in HuggingFace. The models can be requested with the proper permissions from the corresponding authors. 

\section*{Code availability}
To facilitate reproducibility, the source codes can be publicly accessed through github without restrictions at \href{https://github.com/cja5553/LLMs_in_perioperative_care}{https://github.com/cja5553/LLMs\_in\_perioperative\_care}.

\section*{Declarations}

This study is supported, in part by, funding from the Agency for Healthcare Research and Quality (R01 HS029324-02).

\bibliography{sn-bibliography}


\end{document}